\documentclass{article}
\usepackage[preprint]{neurips_2021}
\usepackage[utf8]{inputenc} % allow utf-8 input
\usepackage[T1]{fontenc}    % use 8-bit T1 fonts
\usepackage{hyperref}       % hyperlinks
\usepackage{url}            % simple URL typesetting
\usepackage{booktabs}       % professional-quality tables
\usepackage{amsfonts}       % blackboard math symbols
\usepackage{nicefrac}       % compact symbols for 1/2, etc.
\usepackage{microtype}      % microtypography
\usepackage{xcolor}         % colors
\usepackage{listings}
\usepackage{algorithm}
\usepackage{algorithmicx}
\usepackage{algpseudocode}
\usepackage{multirow}

\usepackage{amsmath,amsfonts,bm}

\def\eqref#1{equation~\ref{#1}}
\def\1{\bm{1}}

\def\va{{\bm{a}}}

\def\vh{{\bm{h}}}

\def\vx{{\bm{x}}}

\DeclareMathAlphabet{\mathsfit}{\encodingdefault}{\sfdefault}{m}{sl}
\SetMathAlphabet{\mathsfit}{bold}{\encodingdefault}{\sfdefault}{bx}{n}

\def\gA{{\mathcal{A}}}

\def\gD{{\mathcal{D}}}

\def\gL{{\mathcal{L}}}

\def\gN{{\mathcal{N}}}

\def\gS{{\mathcal{S}}}

\def\gX{{\mathcal{X}}}

\newcommand{\E}{\mathbb{E}}

\usepackage{subfig}
\usepackage{graphicx}
\usepackage[colorinlistoftodos]{todonotes}
\graphicspath{{./figs/}}
\usepackage[capitalise,nameinlink,noabbrev]{cleveref}

\title{Mastering Visual Continuous Control: \\Improved Data-Augmented Reinforcement Learning}
\author{%
  Denis Yarats \\
  NYU \& FAIR \\
  \And
  Rob Fergus \\
  NYU \\
  \And
  Alessandro Lazaric \\
  FAIR
  \And
  Lerrel Pinto \\
  NYU
  \And
  Code: \url{https://github.com/facebookresearch/drqv2}

}

\newcommand{\drq}{{\bf DrQ-v2}}
\newcommand{\drqs}{DrQ-v2}

\begin{document}

\maketitle
\newif\ifincludeappendix

\begin{abstract}

We present \drqs{}, a model-free reinforcement learning (RL) algorithm for visual continuous control. \drqs{} builds on DrQ, an off-policy actor-critic approach that uses data augmentation to learn directly from pixels. We introduce several improvements that yield state-of-the-art results on the DeepMind Control Suite. Notably, \drqs{} is able to solve complex humanoid locomotion tasks directly from pixel observations, previously unattained by model-free RL. \drqs{} is conceptually simple, easy to implement, and provides significantly better computational footprint compared to prior work, with the majority of tasks taking just $8$ hours to train on a single GPU. Finally, we publicly release~\drqs{}'s implementation to  provide RL practitioners with a strong and computationally efficient baseline.
\end{abstract}

\begin{figure}[b!]
    \centering
    \vspace{-30pt}
    \subfloat[Average performance across $12$ challenging DMC tasks.]{\includegraphics[width=0.7\linewidth]{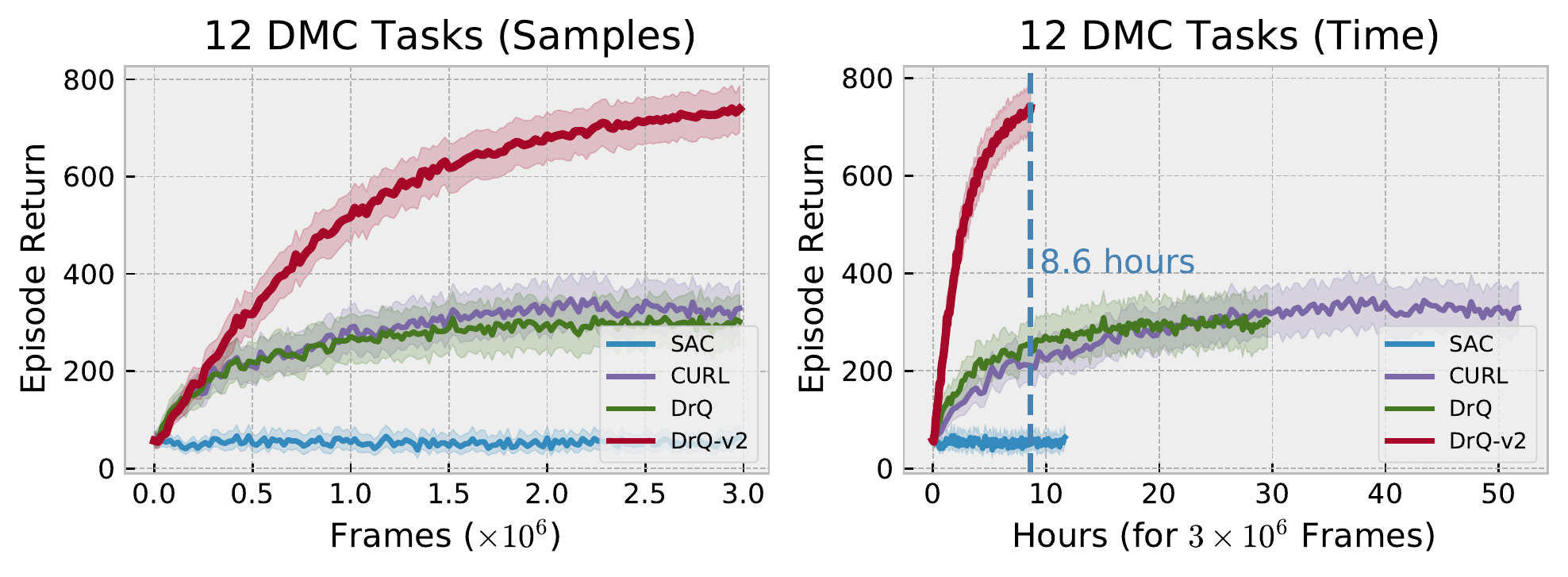} \label{fig:exp_dmc_12}}\\
    \vspace{-10pt}
    \subfloat[Performance on \textit{Humanoid Walk}.]{\includegraphics[width=0.7\linewidth]{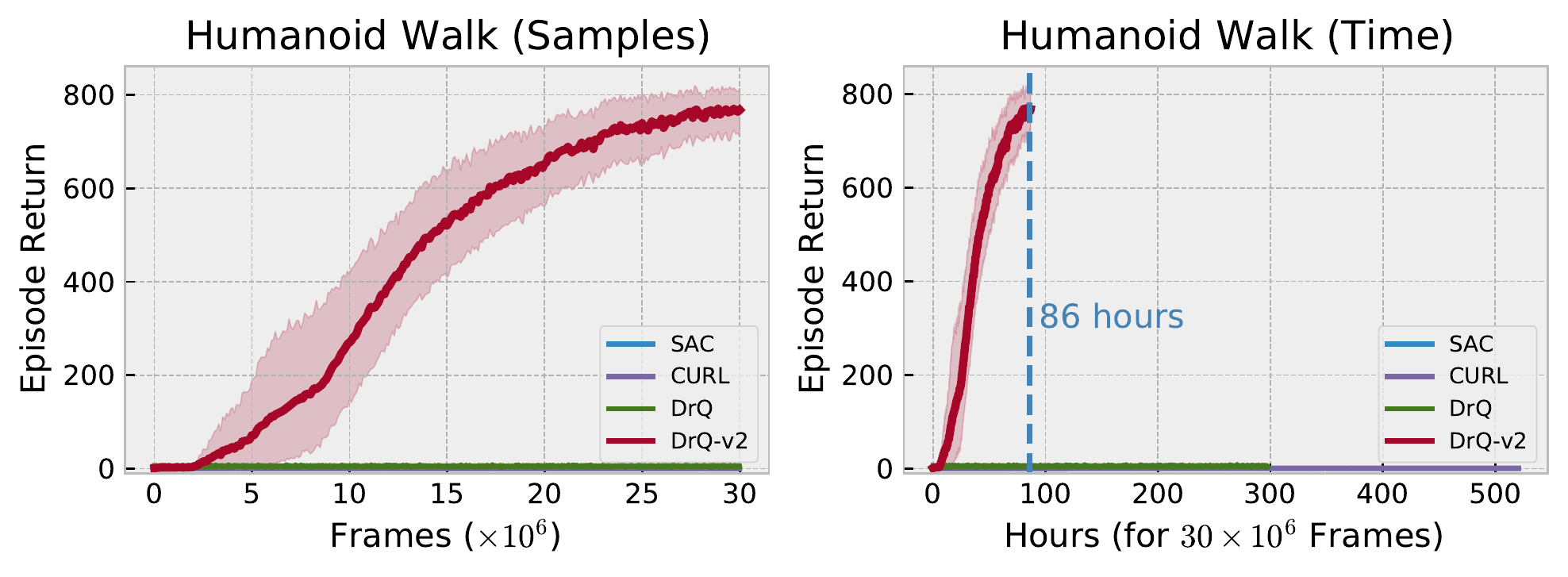} \label{fig:exp_dmc_hum}}
    %\vspace{-10pt}
    \caption{\drq{} demonstrates significantly better sample efficiency and computational footprint compared to state-of-the-art model-free methods for visual continuous control while being conceptually simple and easy to implement. (\textbf{a}) Average performance results across $12$ challenging tasks from the DeepMind Control Suite (the set of tasks can be seen in~\cref{fig:exp_dmc_medium}). (\textbf{b}) Performance on the \textit{Humanoid Walk} task, previously unsolved by model-free methods. In both cases we report sample complexity and wall-clock time axes for evaluation, with time being measured on a single GPU machine and using official implementations for each method. } 
    \label{fig:exp_dmc_main}
    %\vspace{-20pt}
\end{figure}

\section{Introduction}

Creating sample-efficient continuous control methods that observe high-dimensional images has been a long standing challenge in reinforcement learning (RL)~\cite{}. Over the last three years, the RL community has made significant headway on this problem, improving sample-efficiency significantly. The key insight to solving visual control is the learning of better low-dimensional representations, either through autoencoders~\citep{yarats2019improving,finn2015deepspatialae}, variational inference~\citep{hafner2018planet,hafner2019dream,lee2019slac}, contrastive learning~\citep{srinivas2020curl,yarats2021proto}, self-prediction~\citep{schwarzer2020spr}, or data augmentations~\citep{yarats2021image,laskin2020reinforcement}. However, current state-of-the-art model-free methods are still limited in three ways. First, they are unable to solve the more challenging visual control problems such as quadruped and humanoid locomotion. Second, they often require significant computational resources, i.e.~lengthy training times using distributed multi-GPU infrastructure. Lastly, it is often unclear how different design choices affect overall system performance.

\begin{figure}[t!]
    \centering
    % \vspace{-10mm}
    \includegraphics[width=\linewidth]{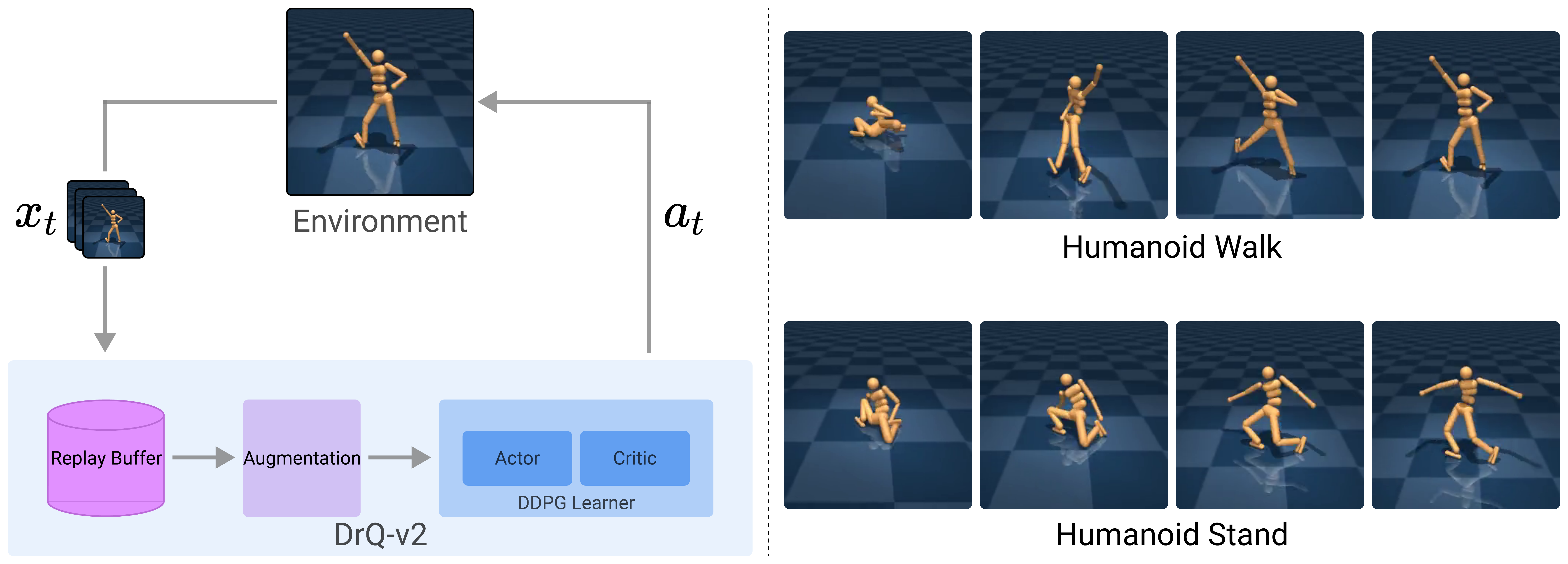}

    \caption{(Left):~\drqs{} is an off-policy actor-critic algorithm for image-based RL. It alleviates encoder overfitting by applying random shift augmentation to pixel observations sampled from the replay buffer. (Right): Examples of walking and standing behaviors learned by~\drqs{} for a complex humanoid agent from DMC~\citep{tassa2018dmcontrol} with $21$ and $54$ dimensional action and state spaces, respectively.~\drqs{} does not have access to the internal state of the environment, only observing three consecutive pixel frames at a time. Despite this imperfect observational channel, our agent still manages to solve the tasks. To the best of our knowledge, this is the first successful demonstration by a model-free method, using pixel-based inputs of this tasks.}
    %of our method is provided in~\cref{ ~minimizes the cross entropy loss to optimize the encoder $f_\theta$, projector $g_\phi$, and discrete prototypical vectors $\{{c_i}\}_{i=1}^M$. The representation $y_t$ obtained from the encoder $f_\theta$ and the prototypes $\{{c_i}\}_{i=1}^M$ are also passed to the RL module to learn a policy and a value function. Importantly, \proto~blocks the gradients from the RL loss function in order to learn \emph{task-agnostic} representations and prototypes, which are only trained with our self supervised loss.}
    \label{fig:model}
    % \vspace{-10pt}
\end{figure}

In this paper we present \drq{}, a simple model-free algorithm that builds on the idea of using data augmentations~\citep{yarats2021image,laskin2020reinforcement} to solve hard visual control problems. Most notably, it is the first model-free method that solves complex humanoid tasks directly from pixels. Compared to previous state-of-the-art model-free methods, \drqs{} provides significant improvements in sample efficiency across tasks from the DeepMind Control Suite~\citep{tassa2018dmcontrol}. Conceptually simple, \drqs{} is also computationally efficient, which allows solving most tasks in DeepMind Control Suite in just $8$ hours on a single GPU (see~\cref{fig:exp_dmc_main}). Recently, a model-based method, DreamerV2~\citep{hafner2020dreamerv2} was also shown to solve visual continuous control problems and it was first to solve the humanoid locomotion problem from pixels. While our model-free \drqs{} matches DreamerV2 in terms sample efficiency and performance, it does so $4\times$ faster in terms of wall-clock time to train. We believe this makes \drqs{} a more accessible approach to support research in visual continuous control and it reinforces the question on whether model-free or model-based is the more suitable approach to solve this type of tasks.

DrQ-v2, which is detailed in~\cref{section:method}, improves upon DrQ~\citep{yarats2021image} by making several algorithmic changes: (i) switching the base RL algorithm from SAC~\citep{haarnoja2018sac} to DDPG~\citep{lillicrap2015ddpg}, (ii) this allows us straightforwardly incorporating multi-step return, (iii) adding bilinear interpolation to the random shift image augmentation, (iv) introducing an exploration schedule, (v) selecting better hyper-parameters including a larger capacity of the replay buffer. A careful ablation study of these design choices is presented in~\cref{section:ablation}. 
Furthermore, we re-examine the original implementation of DrQ and identify several computational bottlenecks such as replay buffer management, data augmentation processing, batch size, and frequency of learning updates (see~\cref{section:code_details}). To remedy these, we have developed a new implementation that both achieves better performance and trains around $3.5$ times faster with respect to wall-clock time than the previous implementation on the same hardware with an increase in environment frame throughput (FPS) from $28$ to $96$ (i.e., it takes $10^6 / 96 / 3600 \approx 2.9$ hours to train for 1M environment steps). %The PyTorch implementation of \drqs{} is available at \url{https://github.com/facebookresearch/drqv2}.

\section{Background}

\subsection{Reinforcement Learning from Images} 
We formulate image-based control as an infinite-horizon Markov Decision Process (MDP)~\citep{bellman1957mdp}. Generally, in such a setting, an image rendering of the system is not sufficient to perfectly describe the system's underlying state. To this end and per common practice~\citep{mnih2013dqn}, we approximate the current state of the system by stacking three consecutive prior observations. With this in mind, such MDP can be described as a tuple $(\gX, \gA, P, R, \gamma, d_0)$, where $\gX$ is the state space (a three-stack of image observations), $\gA$ is the action space, $P: \gX \times \gA \to \Delta(\gX)$ is the transition function\footnote{Here, $\Delta(\gX)$ denotes a distribution over the state space $\gX$.} that defines a probability distribution over the next state given the current state and action, $R: \gX \times \gA \to [0,1]$ is the reward function, $\gamma \in [0, 1)$ is a discount factor, and $d_0 \in \Delta(\gX)$ is the distribution of the initial state $\vx_0$. The goal is to find a policy $\pi: \gX \to \Delta(\gA)$ that maximizes the expected discounted sum of rewards $\E_\pi[\sum_{t=0}^\infty \gamma^t r_t]$, where  $\vx_0 \sim d_0$, and $\forall t$ we have $\va_{t} \sim \pi(\cdot|\vx_{t})$, $\vx_{t+1} \sim P(\cdot| \vx_{t}, \va_{t})$, and $r_t = R(\vx_{t}, \va_{t})$.

\subsection{Deep Deterministic Policy Gradient}
Deep Deterministic Policy Gradient (DDPG)~\citep{lillicrap2015ddpg} is an actor-critic algorithm for continuous control that concurrently learns a Q-function $Q_\theta$ and a deterministic policy $\pi_\phi$. For this, DDPG uses Q-learning~\citep{watkins1992qlearning} to learn $Q_\theta$ by minimizing the one-step Bellman residual 
    $J_\theta(\gD) = \E_{\substack{( \vx_t,\va_t, r_t, \vx_{t+1}) \sim \gD }}[(Q_\theta(\vx_t, \va_t) - r_t - \gamma Q_{\bar{\theta}}(\vx_{t+1},\pi_\phi(\vx_{t+1} ) )^2]$.
The policy $\pi_\phi$ is learned by employing Deterministic Policy Gradient (DPG)~\citep{silver14dpg} and maximizing $J_\phi(\gD) = \E_{\vx_t \sim \gD} [Q_\theta(\vx_t, \pi_\phi(\vx_t))]$, so $\pi_\phi(\vx_t)$ approximates $\mathrm{argmax}_\va Q_\theta(\vx_t, \va)$.  Here, $\gD$ is a replay buffer of environment transitions and $\bar{\theta}$ is an exponential moving average of the weights. DDPG is amenable to incorporate $n$-step returns~\citep{watkins1989phd,peng1996nstepq} when estimating TD error beyond a single step~\citep{barth-maron2018d4pg}. In practice, $n$-step returns allow for faster reward propagation and has been previously used in policy gradient and Q-learning methods~\citep{mnih2016asynchronous,barth-maron2018d4pg,hessel2017rainbow}.

\subsection{Data Augmentation in Reinforcement Learning}
Recently, it has been shown that data augmentation techniques, commonplace in Computer Vision, are also important for achieving the state-of-the-art performance in image-based RL~\citep{yarats2021image,laskin2020reinforcement}. For example, the state-of-the-art algorithm for visual RL, DrQ~\citep{yarats2021image} builds on top of Soft Actor-Critic~\citep{haarnoja2018sac}, a model-free actor-critic algorithm, by adding a convolutional encoder and data augmentation in the form of random shifts. The use of such data augmentations now forms an essential component of several recent visual RL algorithms~\citep{srinivas2020curl,raileanu2020automatic,yarats2021proto,stooke2020decoupling,hansen2021softda,schwarzer2020spr}.

\section{DrQ-v2: Improved Data-Augmented Reinforcement Learning}
\label{section:method}

\begin{algorithm*}[t!]

\caption{
\drqs: Improved data-augmented RL.
}

\begin{algorithmic}
  \State \textbf{Inputs:} \\
    $f_\xi$, $\pi_\phi$, $Q_{\theta_1}$, $Q_{\theta_2}$: parametric networks for encoder, policy, and Q-functions respectively.\\
    $\mathrm{aug}$: random shifts image augmentation.\\
    $\sigma(t)$: scheduled standard deviation for the exploration noise defined in~\cref{eq:decay}. \\
    $T$, $B$, $\alpha$, $\tau$, $c$: training steps, mini-batch size, learning rate, target update rate, clip value.
    \State \textbf{Training routine:}
    \For{each timestep $t=1..T$}
    \State $\sigma_t \leftarrow \sigma(t)$ \Comment{Compute stddev for the exploration noise}
    \State  $\va_t \leftarrow \pi_\phi(f_\xi( \vx_{t})) + \epsilon$ and $\epsilon \sim \gN(0, \sigma_t^2)$ \Comment{Add noise to the deterministic action}
    \State  $\vx_{t+1} \sim P(\cdot | \vx_{t}, \va_t)$ \Comment{Run transition function for one step}
    \State $\mathcal{D} \leftarrow \mathcal{D} \cup (\vx_t, \va_t, R(\vx_t, \va_t), \vx_{t+1})$ 
    \Comment Add a transition to the replay buffer
    \State \textsc{UpdateCritic}($\mathcal{D},\sigma_t$) 
    \State \textsc{UpdateActor}($\mathcal{D},\sigma_t$) 
    
    \EndFor
    \Procedure{UpdateCritic}{$\mathcal{D}, \sigma$}
        \State $\{(\vx_t, \va_t, r_{t:t+n-1}, \vx_{t+n})\} \sim \mathcal{D}$ \Comment Sample a mini batch of $B$ transitions
        \State $\vh_t, \vh_{t+n} \leftarrow f_\xi(\mathrm{aug}(\vx_t)), f_\xi(\mathrm{aug}(\vx_{t+n}))$ \Comment Apply data augmentation and encode
        \State $\va_{t+n} \leftarrow \pi_\phi(\vh_{t+n}) + \epsilon$ and $\epsilon \sim \mathrm{clip}(\gN(0, \sigma^2))$ \Comment{Sample action}
        \State Compute $\gL_{\theta_1, \xi}$ and  $\gL_{\theta_2, \xi}$ using~\cref{eq:critic} \Comment{Compute critic losses}
        \State $\xi \leftarrow \xi -\alpha \nabla_\xi (\gL_{\theta_1, \xi} + \gL_{\theta_2, \xi})  $ \Comment Update encoder weights
        \State $\theta_k \leftarrow \theta_k -\alpha \nabla_{\theta_k} \gL_{\theta_k, \xi} \quad \forall k \in \{1,2\}$ \Comment {Update critic weights}
        \State $\bar{\theta}_k \leftarrow (1-\tau)\bar{\theta}_k + \tau  \theta_k \quad \forall k \in \{1,2\}$ \Comment Update critic target weights
    \EndProcedure
    \Procedure{UpdateActor}{$\mathcal{D},\sigma$}
        \State $\{(\vx_t)\} \sim \mathcal{D}$ \Comment Sample a mini batch of $B$ observations
        \State $\vh_t \leftarrow f_\xi(\mathrm{aug}(\vx_t))$ \Comment Apply data augmentation and encode
        \State $\va_{t} \leftarrow \pi_\phi(\vh_{t}) + \epsilon$ and $\epsilon \sim \mathrm{clip}(\gN(0, \sigma^2))$ \Comment{Sample action}
        \State Compute $\gL_{\phi}$ using~\cref{eq:actor} \Comment{Compute actor loss}
        \State $\phi \leftarrow \phi -\alpha \nabla_\phi \gL_{\phi} $ \Comment Update actor's weights only
    \EndProcedure
\end{algorithmic}

\label{alg:drqv2}
\end{algorithm*}

In this section, we describe~\drq, a simple model-free actor-critic RL algorithm for image-based continuous control, that builds upon DrQ. 

\subsection{Algorithmic Details}
\label{section:algo_details}
\paragraph{Image Augmentation} As in DrQ we apply random shifts image augmentation to pixel observations of the environment. In the settings of visual continuous control by DMC, this augmentation can be instantiated by first padding each side of $84 \times 84$ observation rendering by $4$ pixels (by repeating boundary pixels), and then selecting a random $84 \times 84$ crop, yielding the original image shifted by $\pm 4$ pixels. We also find it useful to apply bilinear interpolation on top of the shifted image (i.e, we replace each pixel value with the average of the four nearest pixel values). In our experience, this modification provides an additional performance boost across the board.
\paragraph{Image Encoder}
The augmented image observation is then embedded into a low-dimensional latent vector by applying a convolutional encoder. We use the same encoder architecture as in DrQ, which first was introduced introduced in SAC-AE~\citep{yarats2019improving}. This process can be succinctly summarized as $\vh = f_\xi(\mathrm{aug}(\vx))$,
where $f_\xi$ is the encoder, $\mathrm{aug}$ is the random shifts augmentation, and $\vx$ is the original image observation.

\paragraph{Actor-Critic Algorithm}
% TODO: explain actor acritic and talk about nxtep return
We use DDPG~\citep{lillicrap2015ddpg} as a backbone actor-critic RL algorithm and, similarly to~\cite{barth-maron2018d4pg}, augment it with $n$-step returns to estimate TD error. This results into faster reward propagation and overall learning progress~\citep{mnih2016a3c}. While some methods~\citep{hafner2020dreamerv2} employ more sophisticated techniques such as TD($\lambda$) or Retrace($\lambda$)~\citep{munos2016retrace}, they are often computationally demanding when $n$ is large. We find that using simple $n$-step returns, without an importance sampling correction, strikes a good balance between performance and efficiency. We also employ clipped double Q-learning~\citep{fujimoto2018td3} to reduce overestimation bias in the target value. Practically, this requires training two Q-functions $Q_{\theta_1}$ and $Q_{\theta_2}$. For this, we sample a mini-batch of transitions $\tau =( \vx_t,\va_t, r_{t:t+n-1}, \vx_{t+n})$ from the replay buffer $\gD$ and compute the following two losses:
\begin{align}
    \gL_{\theta_k, \xi}(\gD) &= \E_{\tau \sim \gD }\big[(Q_{\theta_k}(\vh_t, \va_t) - y) ^2 \big] \quad  \forall k \in \{1,2\},
\label{eq:critic}
\end{align}
with the TD target $y$ defined as:
\begin{align*}
    y &= \sum_{i=0}^{n-1} \gamma^i r_{t+i} + \gamma^n \min_{k=1,2} Q_{\bar{\theta}_k}(\vh_{t+n}, \va_{t+n}),
\end{align*}
where  $\vh_t = f_\xi(\mathrm{aug}(\vx_t))$, $\vh_{t+n}=f_\xi(\mathrm{aug}(\vx_{t+n}))$, $\va_{t+n} = \pi_{\phi}(\vh_{t+n}) + \epsilon$, and $\bar{\theta}_1$,$\bar{\theta}_2$ are the slow-moving weights for the Q target networks. We note, that in contrast to DrQ, we do not employ a target network for the encoder $f_\xi$ and always use the most recent weights $\xi$ to embed $\vx_t$ and $\vx_{t+n}$. The exploration noise $\epsilon$ is sampled from $\mathrm{clip}(\gN(0, \sigma^2), -c, c)$ similar to TD3~\citep{fujimoto2018td3}, with the exception of decaying $\sigma$, which we describe below. Finally, we train the deterministic actor $\pi_\phi$ using DPG with the following loss:
\begin{align}
    \gL_{\phi}(\gD) &= -\E_{\vx_t \sim \gD} \big[\min_{k=1,2} Q_{\theta_k} (\vh_t, \va_t) \big], 
\label{eq:actor}
\end{align}
where  $\vh_{t}=f_\xi(\mathrm{aug}(\vx_{t}))$, $\va_{t} = \pi_{\phi}(\vh_{t}) + \epsilon$, and $\epsilon \sim \mathrm{clip}(\gN(0, \sigma^2), -c, c)$. Similar to DrQ, we do not use actor's gradients to update the encoder's parameters $\xi$.

\paragraph{Scheduled Exploration Noise}
Empirically, we observe that it is helpful to have different levels of exploration at different stages of learning. At the beginning of training we want the agent to be more stochastic and explore the environment more effectively, while at the later stages of training, when the agent has already identified promising behaviors, it is better to be more deterministic and master those behaviors. Similar to~\cite{amos2020modelbased}, we instantiate this idea by using linear decay $\sigma(t)$ for the variance $\sigma^2$ of the exploration noise defined as:
\begin{align}
    \sigma(t) &= \sigma_{\mathrm{init}} + (1 - \min (\frac{t}{T}, 1)) (\sigma_{\mathrm{final}} - \sigma_{\mathrm{init}}),
    \label{eq:decay}
\end{align}
where $\sigma_{\mathrm{init}}$ and  $\sigma_{\mathrm{final}}$ are the initial and final values for standard deviation, and $T$ is the decay horizon.
\paragraph{Key Hyper-Parameter Changes}
We also conduct an extensive hyper-parameter search and identify several useful hyper-parameter modifications compared to DrQ. The three most important hyper-parameters are: (i) the size of the replay buffer, (ii) mini-batch size, and (iii) learning rate. Specifically, we use a $10$ times larger replay buffer than DrQ. We also use a smaller mini-batch size of $256$ without any noticeable performance degradation. This is in contrast to CURL~\citep{srinivas2020curl} and DrQ~\citep{yarats2021image} that both use a larger batch size of $512$ to attain more stable training in the expense of computational efficiency. Finally, we find that using smaller learning rate of $1\times 10^{-4}$, rather than DrQ's learning rate of $1\times 10^{-3}$, results into more stable training without any loss in learning speed.

\subsection{Implementation Details}
\label{section:code_details}

\paragraph{Faster Image Augmentation}

We replace DrQ's random shifts augmentation (i.e., \texttt{kornia.augmentation.RandomCrop}) by a custom implementation that uses flow-field image sampling provided in PyTorch (i.e., \texttt{grid\_sample}). This is done for two reasons. First, we noticed that Kornia's implementation does not fully utilize GPU pipelining since it has some intermediate CPU to GPU data transferring which breaks the computational flow. Second, using \texttt{grid\_sample} allows straightforward addition of bilinear interpolation. Our custom random shifts augmentation improves training throughput by a factor of $2$. 
\paragraph{Faster Replay Buffer}
Another computational bottleneck of DrQ was the replay buffer. The specific implementation had poor memory management which resulted in slow CPU to GPU data transfer, which also restricted the number of image-based transitions that could be stored. We reimplemented the replay buffer to address these issues which led to a ten-fold increase in storage capacity and faster data transfer. More details are available in our open-source release. We note that the improved training speed of~\drqs{} was key to solving humanoid tasks as it enabled much faster experimentation.

\section{Experiments}
In this section we provide empirical evaluation of~\drqs{} on an extensive set of visual continuous control tasks from DMC~\citep{tassa2018dmcontrol}. We first present comparison to prior methods, both model-free and model-based, in terms of sample efficiency and wall-clock time. We then present a large scale ablation study that guided the final version of~\drqs.

\begin{figure}[t!]
    \centering

    \includegraphics[width=\linewidth]{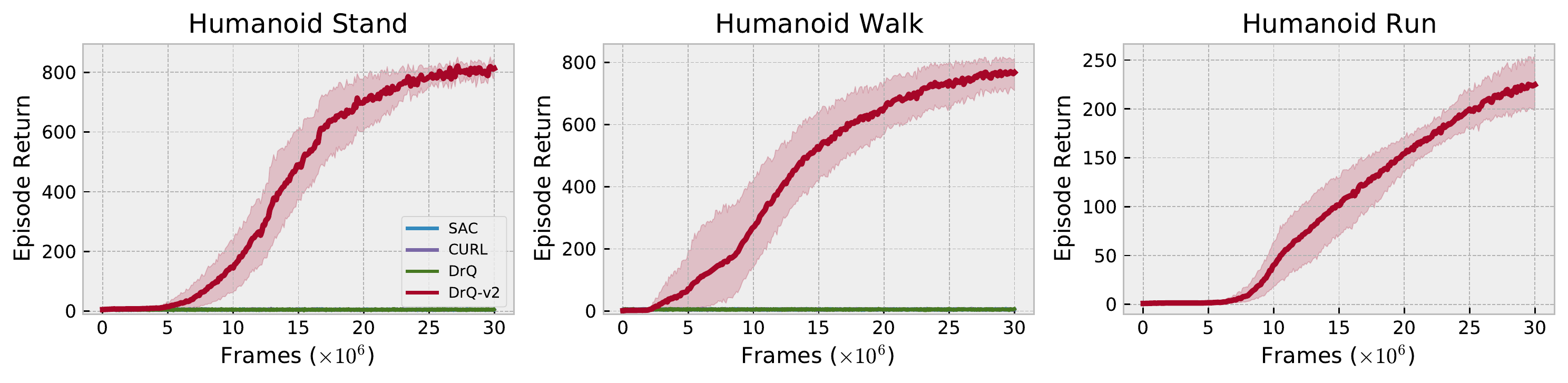}
    \caption{The \textit{hard} benchmark consists of three humanoid locomotion tasks: \textit{stand}, \textit{walk}, and \textit{run}. These three represent particularly hard exploration challenges, being previously unsolvable by model-free methods. The training speed of ~\drqs{} was key to solving the task, since it allowed for extensive investigation of different variations, resulting in the discovery of effective strategies.   }
    \label{fig:exp_dmc_hard}

    % \vspace{-10pt}
\end{figure}

%\subsection{The DeepMind Control Suite Benchmarks}
\subsection{Setup}
\label{section:setup}
\paragraph{Environments}
We consider a set of MuJoCo tasks~\citep{todorov2012mujoco} provided by DMC~\citep{tassa2018dmcontrol}, a widely used benchmark for continous control. DMC offers environments of various difficulty, ranging from the simple control problems such as the single degree of freedom (DOF) pendulum and cartpool, to the control of complex multi-joint bodies such as the humanoid (21 DOF). We consider learning from pixels. In this setting, environment observations are stacks of $3$ consecutive RGB images of size $84 \times 84$, stacked  along the channel dimension to enable inference of dynamic information like velocity and acceleration.  In total, we consider $24$ different tasks, which we group into three buckets, \textit{easy}, \textit{medium}, and \textit{hard}, according to the sample complexity to reach near-optimal performance (see~\cref{section:benchmarks}). Our motivation for this is to encourage RL practitioners to focus on the medium and hard tasks and stop using the easy tasks for evaluation, as they are mostly solved at this point and may no longer provide any valuable signal in comparing different methods.

\paragraph{Training Details}
For all tasks in the suite an episode corresponds to $1000$ steps, where a per-step reward is in the unit interval $[0, 1]$. This upper bounds the episode return to $1000$ making it easier to compute aggregated performance measures across tasks, as we do in~\cref{fig:exp_dmc_12}.
To facilitate fair wall-clock time comparison all algorithms are trained on the same hardware (i.e., a single NVIDIA V100 GPU machine) and evaluated with the same periodicity of $20000$ environment steps. Each evaluation query averages episode returns over $10$ episodes. Per common practice~\citep{hafner2019dream}, we employ action repeat of $2$ and measure sample complexity in the environment steps, rather than the actor steps. In all the figures we plot the mean performance over $10$ seeds together with the shaded regions which represent $95\%$ confidence intervals. A full list of hyper-parameters can be found in~\cref{section:hyperparams}.

\paragraph{Comparison Axes}
In many real-world applications, taking a step in the environment incurs significant computational cost making sample efficiency a critical feature of an RL algorithm. It is hence important to compare RL algorithms in terms of their sample efficiency. We facilitate this comparison by computing an algorithm's performance measured by episode return with respect to environment steps. On the other end, striving low sample complexity often comes at the cost of a poor computational efficiency. Unfortunately, recent deep RL literature has paid very little attention to this important axis which has led to skyrocketing hardware requirements. Such a trend has made it virtually impossible for an RL practitioner with modest hardware capacity to contribute to advancements in image-based RL, leaving research in this area to a few well-equipped labs. To democratize research in visual RL, we additionally propose to compare the agents in terms of wall-clock training time given the same single GPU hardware. We note that it is possible to adapt~\drqs{} to a distributed setup, as has been done for DDPG in prior work~\citep{barth-maron2018d4pg,hoffman2020acme}.

\subsection{Comparison to Model-Free Methods}

\paragraph{Baselines}%\todo{This section overlaps a bit with 4.1. I would bring most of the general definitions from 4.2 to 4.1 and keep here only comments on the results.}
 We compare our method to several state-of-the-art model-free algorithms for visual RL including CURL~\citep{srinivas2020curl}, DrQ~\citep{yarats2021image}, and vanilla SAC~\citep{haarnoja2018sac} augmented with the convolutional encoder from SAC-AE~\citep{yarats2019improving}. Vanilla SAC is a weak baseline and only included as a ground point to showcase the recent progress in visual RL.

\begin{figure}[t!]
    \centering

    \includegraphics[width=\linewidth]{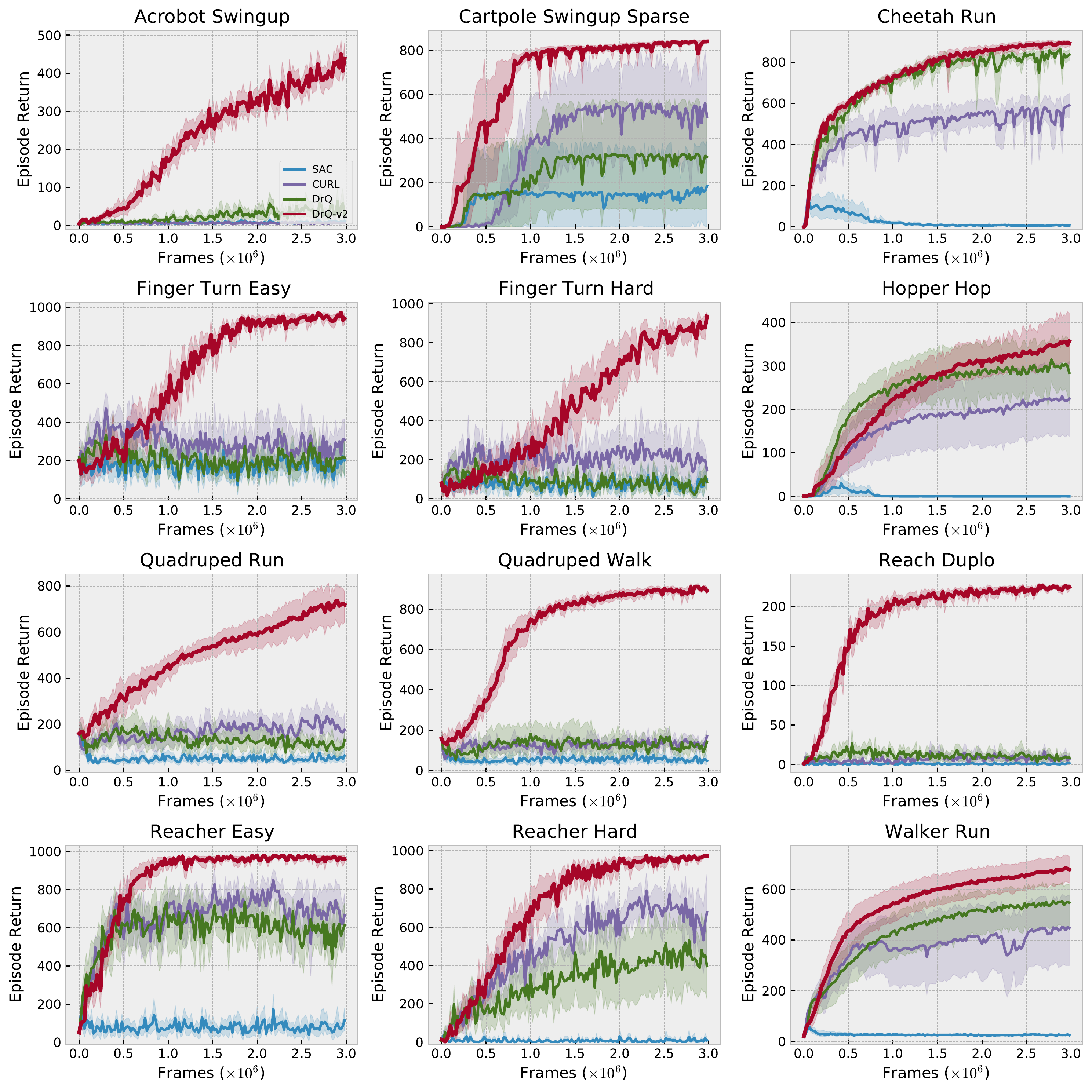}
    \caption{The \textit{medium} benchmark consists of $12$ complex control tasks that offer various challenges, including complex dynamics, sparse rewards, hard exploration, and more.~\drqs{} demonstrates favorable sample efficiency and comfortably outperforms leading model-free baselines.   }
    \label{fig:exp_dmc_medium}
    % \vspace{-10pt}
\end{figure}

\begin{figure}[t!]
    \centering

    \includegraphics[width=\linewidth]{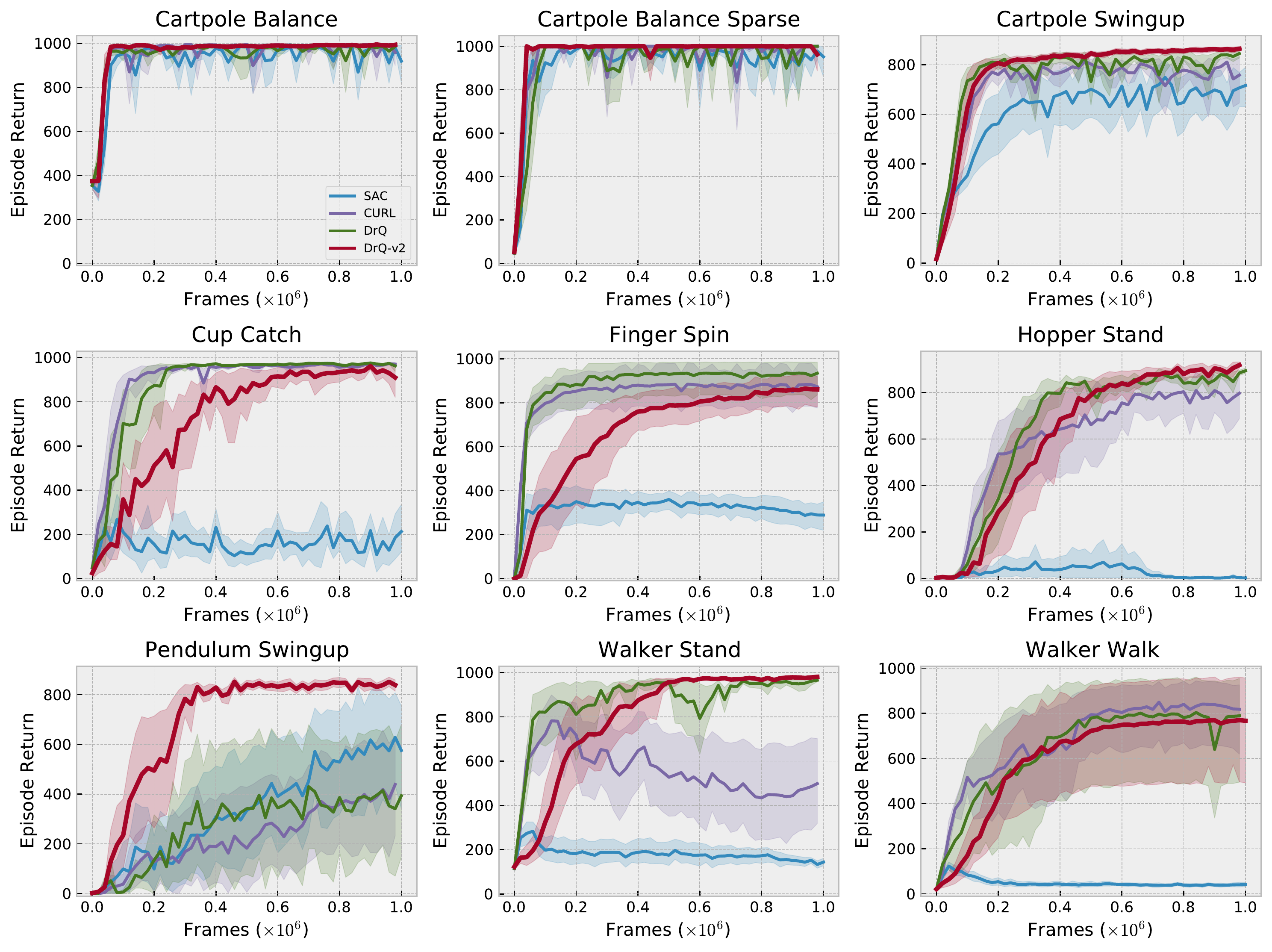}
    \caption{The \textit{easy} benchmark consists of $9$ tasks, where performance gains have been largely saturated by prior work. Still,~\drqs{} is able to match sample complexity of the baselines. We note, that evaluation on these tasks is done for completeness reasons only, and encourage RL practitioners to refrain from using them for benchmarking purposes in future research.  }
    \label{fig:exp_dmc_easy}
    % \vspace{-10pt}
\end{figure}

\paragraph{Sample Efficiency Axis}
We present results on the hard (\cref{fig:exp_dmc_hard}), medium (\cref{fig:exp_dmc_medium}), and easy (\cref{fig:exp_dmc_easy}) subsets of the DMC tasks, where the maximum number of environment interactions is limited to \textit{thirty}, \textit{three}, and \textit{one}  million of steps respectively.Our empirical study reveals that~\drqs{} outperforms prior model-free methods in terms of sample efficiency across the three benchmarks with different levels of difficulty. Importantly,~\drqs{}'s advantage is more pronounced on harder tasks (i.e., acrobot, quadruped, and humanoid), where exploration is especially challenging. Finally,~\drqs{} solves the DMC humanoid locomotion tasks directly from pixels, which, to the best of our knowledge, is the first successful demonstration of such feat by a model-free method.

\begin{figure}[t!]
    \centering

    \includegraphics[width=\linewidth]{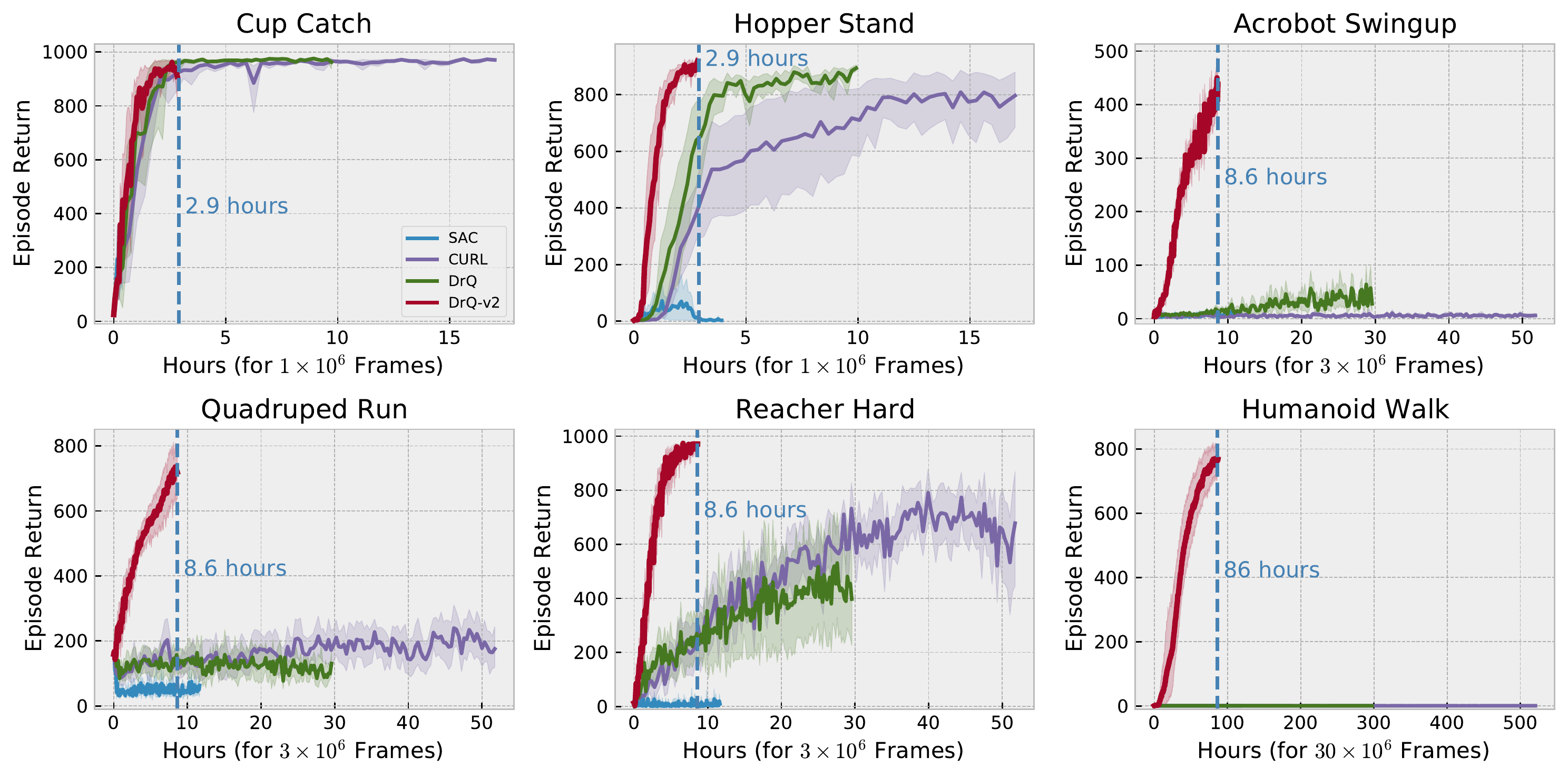}
    \caption{\drqs{} not only achieves superior sample efficiency than prior model-free methods, but it also requires less wall-clock training time to do so. Our benchmarking shows that~\drqs{} can reach a throughput of $96$ FPS on a single NVIDIA V100 GPU. Practically, this means that most of the task can be solved in $8$ hours or less, which greatly speeds up the research process.}
    \label{fig:exp_wall_clock}
    % \vspace{-10pt}
\end{figure}

\paragraph{Compute Efficiency Axis}
To facilitate a fair comparison in terms of sheer wall-clock training time, besides employee the identical training protocol (see~\cref{section:setup}), we also use the same mini-batch size of $256$ for each agent. 
In~\cref{fig:exp_wall_clock}, we evaluate~\drqs{} on a subset of DMC tasks for the sake of brevity only, and note that the demonstrated results can be easily extrapolated to the other tasks given the linear dependency between training time and sample complexity. In our benchmarks,~\drqs{} is able to achieve a throughput of $96$ FPS, which favorably compares to DrQ's $28$ FPS (a $3.4\times$ increase), and CURL's $16$ FPS (a $6\times$ increase) throughputs. Practically,~\drqs{} solves easy, medium, and hard tasks within $2.9$, $8.6$, and $86$ hours respectively. 

% We hope that such training speed would make~\drqs{} a go-to backbone algorithm for further research in imaged-based RL.

\subsection{Comparison to Model-Based Methods}

\paragraph{Baseline}
To see how~\drqs{} stacks up against model-based methods, which tend to achieve better sample complexity in expense of a larger computational footprint, we also compare to recent and unpublished\footnote{ArXiv v3 revision from May 3, 2021 introduces a new result on the \textit{Humanoid Walk} task in Appendix A.} improvements to Dreamer-v2~\citep{hafner2020dreamerv2}, a leading model-based approach for visual continuous control. The recent update shows that the model-based approach can solve the DMC humanoid tasks directly from pixel inputs. The open-source implementation of Dreamer-v2 (\url{https://github.com/danijar/dreamerv2}) only provides learning curves for \textit{Humanoid Walk}. For this reason we run their code to obtain results on other DMC tasks. To limit hardware requirements of compute-expensive Dreamer-v2, we only run it on a subset of $12$ out of $24$ considered tasks. This subset, however, overlaps with all the three (i.e. easy, medium, and hard) benchmarks.

\paragraph{Sample Efficiency Axis}

\begin{figure}[t!]
    \centering

    \includegraphics[width=\linewidth]{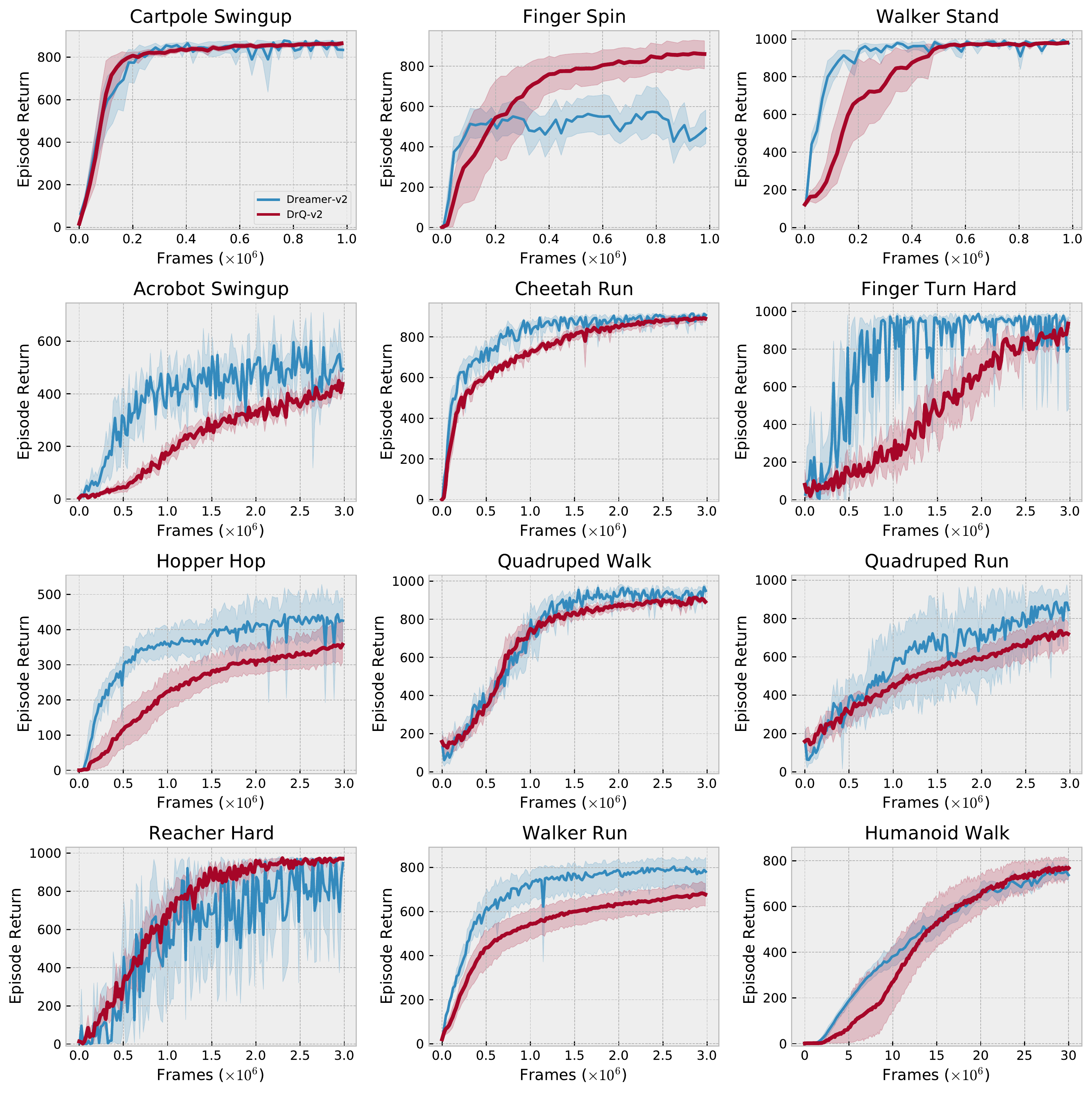}
    \caption{In many cases model-free~\drqs{} can rival model-based Dreamer-v2 in sample efficiency. There are several tasks, however, where Dreamer-v2 performs better. }
    \label{fig:exp_dmc_dreamer}

    % \vspace{-10pt}
\end{figure}

Our empirical study in~\cref{fig:exp_dmc_dreamer} reveals that in many cases,~\drqs{}, despite being a model-free method, can rival sample efficiency of state-of-the-art model-based Dreamer-v2. We note, however, that on several tasks (for example \textit{Acrobot Swingup} and \textit{Finger Turn Hard}) Dreamer-v2 outperforms~\drqs{}. We leave investigation of such discrepancy for future work.

\begin{figure}[t!]
    \centering

    \includegraphics[width=\linewidth]{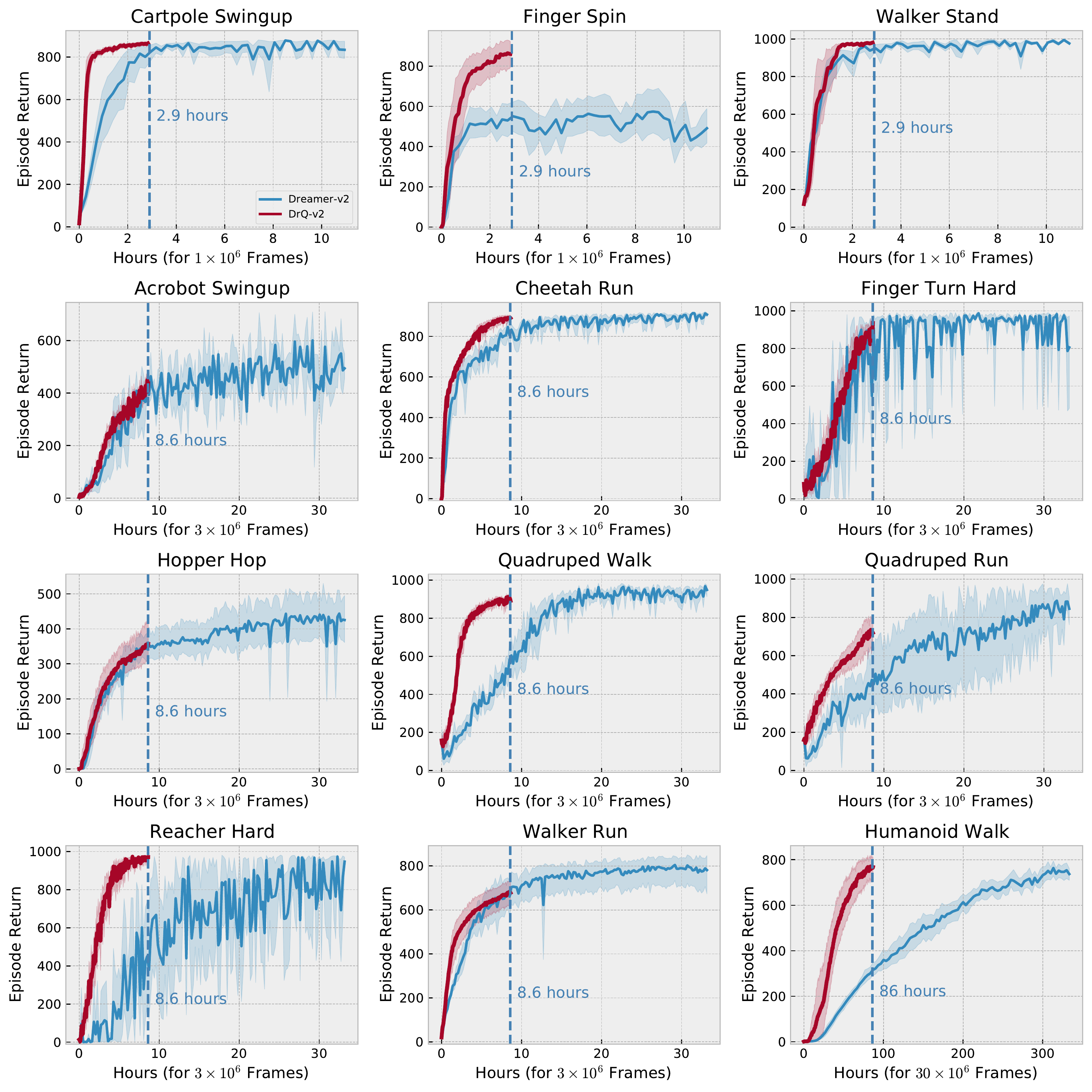}
    \caption{Model-based Dreamer-v2 performs more computations  than model-free~\drqs{}. This allows~\drqs{} to train faster in terms of wall-clock time and outperform Dreamer-v2 in this aspect.}
    \label{fig:exp_dmc_dreamer_wall_clock}
    % \vspace{-10pt}
\end{figure}

\paragraph{Compute Efficiency Axis} A different picture emerges if comparison is done with respect to wall-clock training time. Dreamer-v2, being a model-based method, performs  significantly more floating point operations to reach its sample efficiency. In our benchmarks, Dreamer-v2 records a throughput of $24$ FPS, which is $4\times$ less than~\drqs{}'s throughput of $96$ FPS, measured on the same hardware. In~\cref{fig:exp_dmc_dreamer_wall_clock} we plot learning curves against wall-clock time and observe that~\drqs{} takes less time to solve the tasks.

\subsection{Ablation Study}
\label{section:ablation}
\begin{figure*}[t!]
    \centering
    
    \subfloat[DrQ (dotted silver) relies on SAC as a base RL algorithm. Replacing SAC with DDPG results in a significant performance gain (blue). ]{\includegraphics[width=\linewidth]{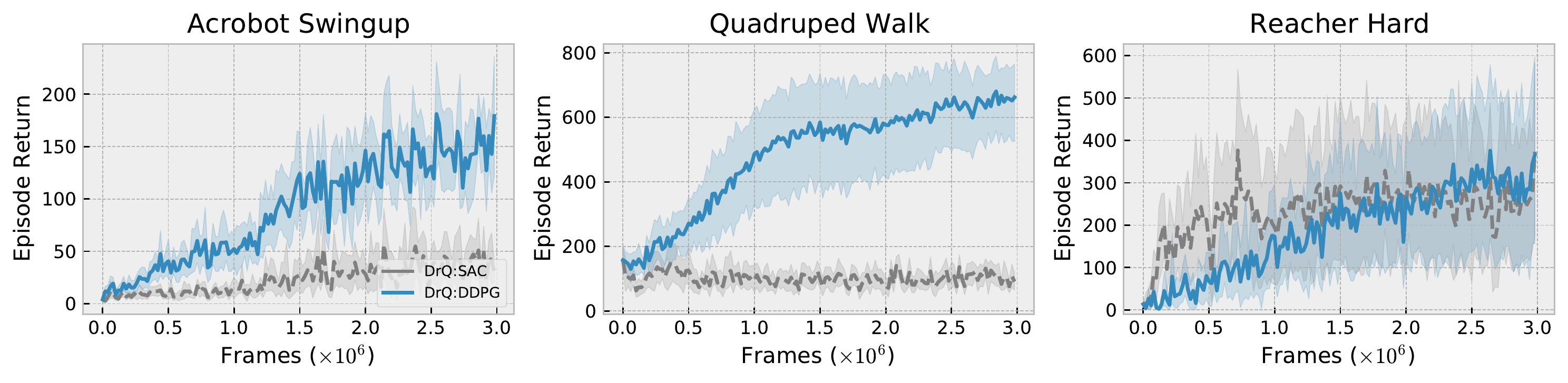}  \label{fig:exp_sac_vs_ddpg}}\\
 
    \subfloat[DDPG straightforwardly incorporates $n$-step returns, a critical tool for exploration. We observe that the $3$ (blue) and $5$ (red) steps variants provide additional improvements to the previous version that uses single step TD-targets (silver). Going forward, we adopt $3$-step returns (blue).]{\includegraphics[width=\linewidth]{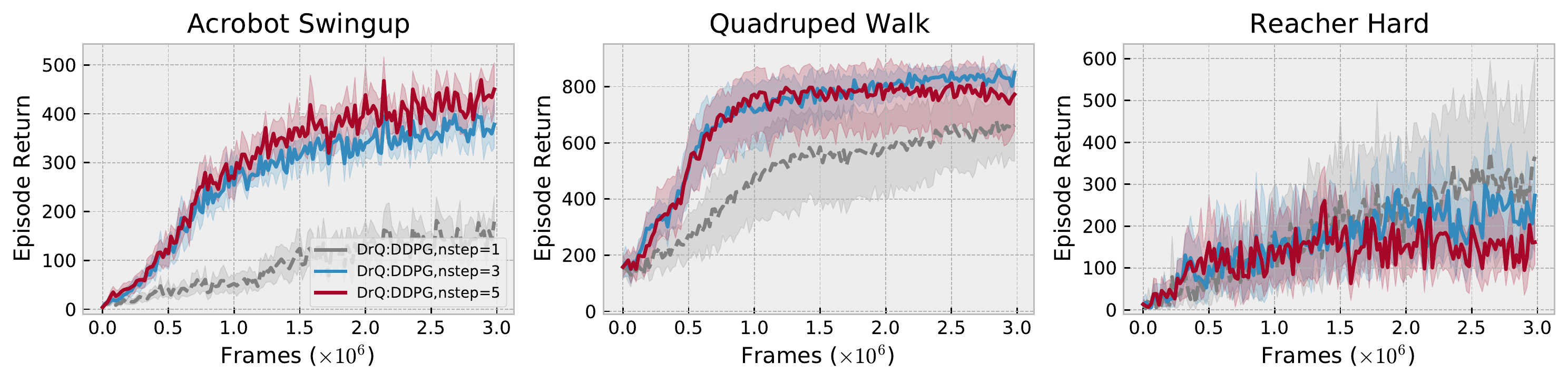}  \label{fig:exp_nstep}}\\
    
    \subfloat[Increasing the size of the replay buffer improves performance, over the original 100k used by DrQ (silver). Going forward, we use a buffer size of 1M (red). ]{\includegraphics[width=\linewidth]{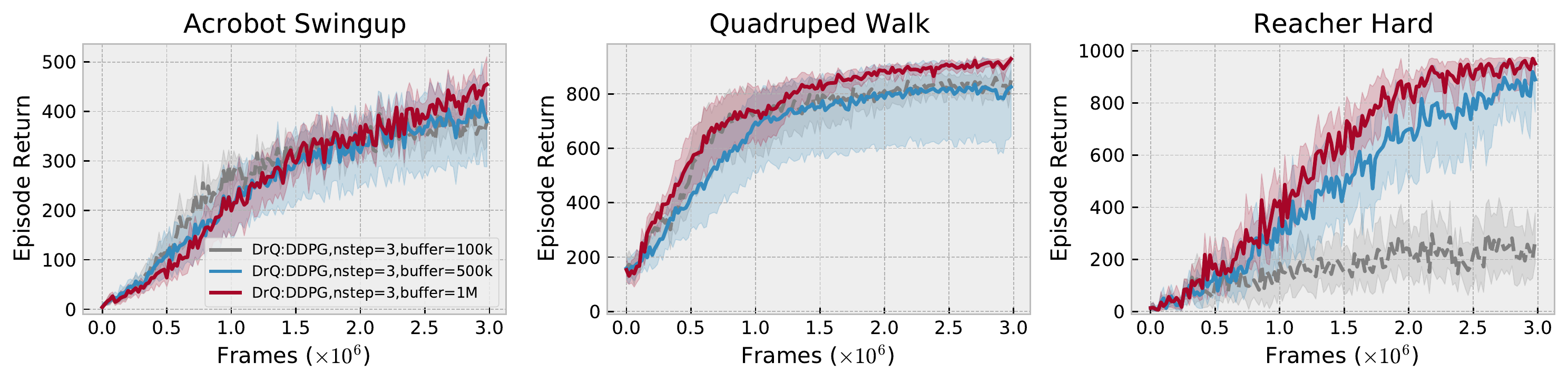}  \label{fig:exp_buffer}}\\

    \subfloat[Finally, a decaying schedule for the variance of the exploration noise (blue) helps on hard exploration tasks, versus the fixed variance variant (silver).]{\includegraphics[width=\linewidth]{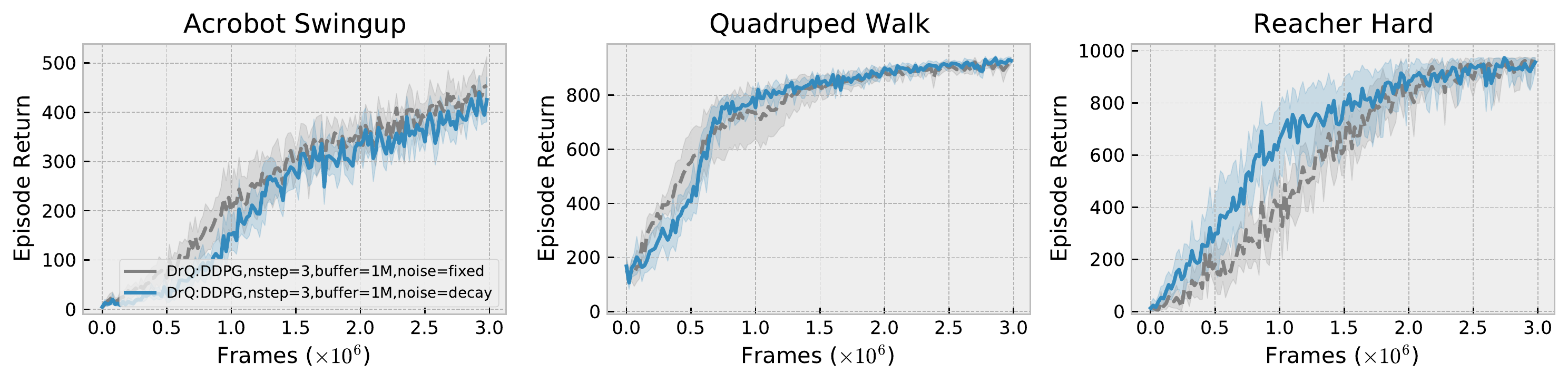}  \label{fig:exp_noise}}\\

         \caption{An ablation study that led us to the final version of~\drqs{}. We incrementally show each of the four key improvements to DrQ that collectively form~\drqs{}. The silver dotted curves in the first row show the original DrQ. In subsequent rows they show progressive improvements, using the optimal choice from the previous rows (i.e., the silver curve in the third row shows DrQ with a DDPG base RL algorithm and $3$-step returns). The red and blue curves show the effect of individual modifications. In the last row the blue curve corresponds to~\drqs{}.}

    \label{fig:ablation}
    % \vspace{-10pt}
\end{figure*}

In this section we present an extensive ablation study that guided us to the final version of~\drqs{}. Here, for brevity we only discuss experiments that were most impactful and omit others that did not pan out. For computational reasons, we only ablate on $3$ different control tasks of various difficulty levels. Our findings are summarized in~\cref{fig:ablation} and detailed below.

\paragraph{Switching from SAC to DDPG} DrQ~\citep{yarats2021image} leverages SAC~\citep{haarnoja2018sac} as the backbone RL algorithm. While it has been demonstrated by many works, including the original manuscripts~\citep{haarnoja2018sac,haarnoja2018sacapplication} that SAC is superior to DDPG~\citep{lillicrap2015continuous}, our careful examination identifies two shortcomings that preclude SAC (within DrQ) to solve hard exploration-wise image-based tasks. First, the automatic entropy adjustment strategy, introduced in~\cite{haarnoja2018sacapplication}, is inadequate and in some cases leads to a premature entropy collapse. This prevents the agent from finding more optimal behaviors due to the insufficient exploration. In~\cref{fig:exp_sac_vs_ddpg}, we empirically verify our intuition and, indeed, observe that DDPG demonstrates better exploration properties than SAC. Here, DDPG uses constant $\sigma=0.2$ for the exploration noise.

\paragraph{$\mathbf{N}$-step Returns}
The second issue concerns the inability of soft Q-learning to incorporate $n$-step returns to estimate TD error in a straightforward manner. The reason for this is that computing a target value for soft Q-function requires estimating per-step entropy of the policy, which is challenging to do for large $n$ in the off-policy regime. In contrast, DDPG does not require estimating per-step entropy to compute targets and is more amenable for $n$-step returns. In~\cref{fig:exp_nstep} we demonstrate that estimating TD error with $n$-step returns improves sample efficiency over vanilla DDPG. We select $3$-step returns as a sensible choice for our method.

%We empirically verify our intuition in~\cref{fig:exp_sac_vs_ddpg}.  we experiment with DDPG that is more amenable for $n$-step returns~\citep{barth-maron2018d4pg}.%\todo{Why is importance sampling more needed in SAC than in DDPG?}
%Furthermore, we use simple isotropic zero-mean Gaussian noise around the policy to facilitate exploration to replace SAC's automatic entropy adjustment. In~\cref{fig:exp_sac_vs_ddpg} we observe that DrQ with vanilla DDPG already improves performance.% \todo{Most of these things were mentioned already in 3.2. It clarifies better the n-step part though.}

\paragraph{Replay Buffer Size}
We hypothesize that a larger replay buffer plays an important role in circumventing the catastrophic forgetting problem~\citep{fedus2020replay}. This issue is especially prominent in tasks with more diverse initial state distributions (i.e., reacher or humanoid tasks), where the vast variety of possible behaviors requires significantly larger memory. We confirm this intuition by ablating the size of the replay buffer in~\cref{fig:exp_buffer}, where we observe that a buffer size of 1M helps to improve performance on \textit{Reacher Hard} considerably.

\paragraph{Scheduled Exploration Noise}
Finally, we demonstrate that it is useful to decay the variance of the exploration noise over the course of training according to~\cref{eq:decay}. In~\cref{fig:exp_noise}, we compare two versions of our algorithm, where the first variant uses a fixed standard deviation of $\sigma=0.2$, while the second variant employes the decaying schedule $\sigma(t)$, with parameters $\sigma_{\mathrm{init}}=1.0$, $\sigma_{\mathrm{final}}=0.1$, and $T=500000$. Having the exploration noise to decay linearly over time turns out to be helpful and provide an additional performance boost, which was especially useful for solving humanoid tasks.

\section{Related Work}

% \subsection{Visual Reinforcement Learning}
\paragraph{Visual Reinforcement Learning} Successes of visual representation learning in computer vision~\citep{vincent2008extracting, doersch2015unsupervised,Wang_UnsupICCV2015, noroozi2016unsupervised,zhang2017split,gidaris2018unsupervised} has inspired successes in visual RL, where coherent representations are learned alongside RL. Works such as SAC-AE~\citep{yarats2019improving}, PlaNet~\citep{hafner2018planet}, and SLAC~\citep{lee2019stochastic}, demonstrated how auto-encoders~\citep{finn2015deepspatialae} could improve visual RL. Following this, other self-supervised objectives such as contrastive learning in CURL~\citep{srinivas2020curl} and ATC~\citep{stooke2020decoupling}, self-prediction in SPR~\citep{schwarzer2020data}, contrastive cluster assignment in Proto-RL~\citep{yarats2021proto}, and augmented data in DrQ~\citep{yarats2021image} and RAD~\citep{laskin2020reinforcement}, have significantly bridged the gap between state-based and image-based RL. Future prediction objectives~\citep{hafner2018planet,hafner2019dream,yan2020learning,finn2015deepspatialae,pinto2016curious,agrawal2016learning} and other auxiliary objectives~\citep{jaderberg2016reinforcement,zhan2020framework,young2020visual,chen2020robust} have shown improvements on a variety of problems ranging from gameplay, continuous control, and robotics. In the context of visual control settings, clever use of augmented data~\citep{yarats2021image,laskin2020reinforcement} currently produces state-of-the-art results on visual tasks from DMC~\citep{tassa2018dmcontrol}. %This paper improves on DrQ~\citep{yarats2021image} by introducing several key modifications, and yields an algorithm 

\paragraph{Humanoid Control}  The humanoid control problem first presented in~\cite{tassa2012trajopt}, has been studied as one of the hardest control problems due to its large state and action spaces. The earliest solutions to this problem use ideas in model-based optimal control to generate policies given an accurate model of the humanoid~\cite{}. Subsequent works in RL have shown that model-free policies can solve the humanoid control problem given access to proprioceptive state observations. However, solving such a problem from visual observations has been a challenging problem, with leading RL algorithms making little progress to solve the task~\citep{tassa2018dmcontrol}. Recently, \cite{hafner2020dreamerv2} was able to solve this problem through a model-based technique in around 30M environment steps and $340$ hours of training on a single GPU machine. \drqs{}, presented in this paper, marks the first model-free RL method that can solve humanoid control from visual observations, taking also around 30M steps and $86$ hours of training on the same hardware.

\section{Conclusion}

We have introduced a conceptually simple model-free actor-critic RL algorithm for image-based continuous control -- \drqs{}. Our method provides significantly better computational footprint and masters tasks from DMC~\citep{tassa2018dmcontrol} directly from pixels, most notably the humanoid locomotion tasks that were previously unsolved by model-free approaches. Additionally, we have provided an efficient PyTorch implementation of \drqs{} that is publicly available at \url{https://github.com/facebookresearch/drqv2}. We hope that our algorithm will help to inspire and democratize further research in visual RL.

\section{Acknowledgements}
This research is supported in part by DARPA through the
Machine Common Sense Program. We thank Brandon Amos, Kimin Lee, Mandi Zhao, and Younggyo Seo for insightful discussions that helped to shape our paper.

\newpage 

\bibliography{main}

\begin{thebibliography}{51}
\providecommand{\natexlab}[1]{#1}
\providecommand{\url}[1]{\texttt{#1}}
\expandafter\ifx\csname urlstyle\endcsname\relax
  \providecommand{\doi}[1]{doi: #1}\else
  \providecommand{\doi}{doi: \begingroup \urlstyle{rm}\Url}\fi

\bibitem[Agrawal et~al.(2016)Agrawal, Nair, Abbeel, Malik, and
  Levine]{agrawal2016learning}
Pulkit Agrawal, Ashvin Nair, Pieter Abbeel, Jitendra Malik, and Sergey Levine.
\newblock Learning to poke by poking: experiential learning of intuitive
  physics.
\newblock In \emph{Proceedings of the 30th International Conference on Neural
  Information Processing Systems}, pages 5092--5100, 2016.

\bibitem[Amos et~al.(2020)Amos, Stanton, Yarats, and
  Wilson]{amos2020modelbased}
Brandon Amos, Samuel Stanton, Denis Yarats, and Andrew~Gordon Wilson.
\newblock On the model-based stochastic value gradient for continuous
  reinforcement learning.
\newblock \emph{CoRR}, 2020.

\bibitem[Barth-Maron et~al.(2018)Barth-Maron, Hoffman, Budden, Dabney, Horgan,
  TB, Muldal, Heess, and Lillicrap]{barth-maron2018d4pg}
Gabriel Barth-Maron, Matthew~W. Hoffman, David Budden, Will Dabney, Dan Horgan,
  Dhruva TB, Alistair Muldal, Nicolas Heess, and Timothy Lillicrap.
\newblock Distributional policy gradients.
\newblock In \emph{International Conference on Learning Representations}, 2018.

\bibitem[Bellman(1957)]{bellman1957mdp}
Richard Bellman.
\newblock A markovian decision process.
\newblock \emph{Indiana Univ. Math. J.}, 1957.

\bibitem[Chen et~al.(2020)Chen, Sax, Lewis, Armeni, Savarese, Zamir, Malik, and
  Pinto]{chen2020robust}
Bryan Chen, Alexander Sax, Gene Lewis, Iro Armeni, Silvio Savarese, Amir~Roshan
  Zamir, Jitendra Malik, and Lerrel Pinto.
\newblock Robust policies via mid-level visual representations: An experimental
  study in manipulation and navigation.
\newblock \emph{CoRR}, 2020.

\bibitem[Doersch et~al.(2015)Doersch, Gupta, and
  Efros]{doersch2015unsupervised}
Carl Doersch, Abhinav Gupta, and Alexei~A Efros.
\newblock Unsupervised visual representation learning by context prediction.
\newblock In \emph{Proceedings of the IEEE International Conference on Computer
  Vision}, pages 1422--1430, 2015.

\bibitem[eng and Williams(1996)]{peng1996nstepq}
Jing eng and Ronald~J. Williams.
\newblock Incremental multi-step q-learning.
\newblock \emph{Machine Learning}, 1996.

\bibitem[Fedus et~al.(2020)Fedus, Ramachandran, Agarwal, Bengio, Larochelle,
  Rowland, and Dabney]{fedus2020replay}
William Fedus, Prajit Ramachandran, Rishabh Agarwal, Yoshua Bengio, Hugo
  Larochelle, Mark Rowland, and Will Dabney.
\newblock Revisiting fundamentals of experience replay.
\newblock \emph{CoRR}, 2020.

\bibitem[Finn et~al.(2015)Finn, Tan, Duan, Darrell, Levine, and
  Abbeel]{finn2015deepspatialae}
Chelsea Finn, Xin~Yu Tan, Yan Duan, Trevor Darrell, Sergey Levine, and Pieter
  Abbeel.
\newblock Learning visual feature spaces for robotic manipulation with deep
  spatial autoencoders.
\newblock \emph{CoRR}, 2015.

\bibitem[Fujimoto et~al.(2018)Fujimoto, van Hoof, and Meger]{fujimoto2018td3}
Scott Fujimoto, Herke van Hoof, and David Meger.
\newblock Addressing function approximation error in actor-critic methods.
\newblock In \emph{Proceedings of the 35th International Conference on Machine
  Learning, {ICML} 2018, Stockholmsmassan, Stockholm, Sweden, July 10-15,
  2018}, 2018.

\bibitem[Gidaris et~al.(2018)Gidaris, Singh, and
  Komodakis]{gidaris2018unsupervised}
Spyros Gidaris, Praveer Singh, and Nikos Komodakis.
\newblock Unsupervised representation learning by predicting image rotations,
  2018.

\bibitem[Haarnoja et~al.(2018{\natexlab{a}})Haarnoja, Zhou, Hartikainen,
  Tucker, Ha, Tan, Kumar, Zhu, a~nd Pieter~Abbeel, and
  Levine]{haarnoja2018sacapplication}
Tuomas Haarnoja, Aurick Zhou, Kristian Hartikainen, George Tucker, Sehoon Ha,
  Jie Tan, Vikash Kumar, Henry Zhu, Abhishek~Gupta a~nd Pieter~Abbeel, and
  Sergey Levine.
\newblock Soft actor-critic algorithms and applications.
\newblock \emph{CoRR}, 2018{\natexlab{a}}.

\bibitem[Haarnoja et~al.(2018{\natexlab{b}})Haarnoja, Zhou, Hartikainen,
  Tucker, Ha, Tan, Kumar, Zhu, Gupta, Abbeel, et~al.]{haarnoja2018sac}
Tuomas Haarnoja, Aurick Zhou, Kristian Hartikainen, George Tucker, Sehoon Ha,
  Jie Tan, Vikash Kumar, Henry Zhu, Abhishek Gupta, Pieter Abbeel, et~al.
\newblock Soft actor-critic algorithms and applications.
\newblock \emph{arXiv preprint arXiv:1812.05905}, 2018{\natexlab{b}}.

\bibitem[Hafner et~al.(2018)Hafner, Lillicrap, Fischer, Villegas, Ha, Lee, and
  Davidson]{hafner2018planet}
Danijar Hafner, Timothy Lillicrap, Ian Fischer, Ruben Villegas, David Ha,
  Honglak Lee, and James Davidson.
\newblock Learning latent dynamics for planning from pixels.
\newblock \emph{arXiv preprint arXiv:1811.04551}, 2018.

\bibitem[Hafner et~al.(2019)Hafner, Lillicrap, Ba, and
  Norouzi]{hafner2019dream}
Danijar Hafner, Timothy Lillicrap, Jimmy Ba, and Mohammad Norouzi.
\newblock Dream to control: Learning behaviors by latent imagination.
\newblock \emph{arXiv preprint arXiv:1912.01603}, 2019.

\bibitem[Hafner et~al.(2020)Hafner, Lillicrap, Norouzi, and
  Ba]{hafner2020dreamerv2}
Danijar Hafner, Timothy~P. Lillicrap, Mohammad Norouzi, and Jimmy Ba.
\newblock Mastering atari with discrete world models.
\newblock \emph{CoRR}, 2020.

\bibitem[Hansen and Wang(2021)]{hansen2021softda}
Nicklas Hansen and Xiaolong Wang.
\newblock Generalization in reinforcement learning by soft data augmentation.
\newblock In \emph{International Conference on Robotics and Automation}, 2021.

\bibitem[Hessel et~al.(2017)Hessel, Modayil, van Hasselt, Schaul, Ostrovski,
  Dabney, Horgan, Piot, Azar, and Silver]{hessel2017rainbow}
Matteo Hessel, Joseph Modayil, Hado van Hasselt, Tom Schaul, Georg Ostrovski,
  Will Dabney, Daniel Horgan, Bilal Piot, Mohammad~Gheshlaghi Azar, and David
  Silver.
\newblock Rainbow: Combining improvements in deep reinforcement learning.
\newblock \emph{arXiv preprint arXiv:1710.02298}, 2017.

\bibitem[Hoffman et~al.(2020)Hoffman, Shahriari, Aslanides, Barth-Maron,
  Behbahani, Norman, Abdolmaleki, Cassirer, Yang, Baumli, Henderson, Novikov,
  Colmenarejo, Cabi, Gulcehre, Paine, Cowie, Wang, Piot, and
  de~Freitas]{hoffman2020acme}
Matt Hoffman, Bobak Shahriari, John Aslanides, Gabriel Barth-Maron, Feryal
  Behbahani, Tamara Norman, Abbas Abdolmaleki, Albin Cassirer, Fan Yang, Kate
  Baumli, Sarah Henderson, Alex Novikov, Sergio~Gómez Colmenarejo, Serkan
  Cabi, Caglar Gulcehre, Tom~Le Paine, Andrew Cowie, Ziyu Wang, Bilal Piot, and
  Nando de~Freitas.
\newblock Acme: A research framework for distributed reinforcement learning,
  2020.

\bibitem[Jaderberg et~al.(2016)Jaderberg, Mnih, Czarnecki, Schaul, Leibo,
  Silver, and Kavukcuoglu]{jaderberg2016reinforcement}
Max Jaderberg, Volodymyr Mnih, Wojciech~Marian Czarnecki, Tom Schaul, Joel~Z
  Leibo, David Silver, and Koray Kavukcuoglu.
\newblock Reinforcement learning with unsupervised auxiliary tasks, 2016.

\bibitem[Laskin et~al.(2020)Laskin, Lee, Stooke, Pinto, Abbeel, and
  Srinivas]{laskin2020reinforcement}
Michael Laskin, Kimin Lee, Adam Stooke, Lerrel Pinto, Pieter Abbeel, and
  Aravind Srinivas.
\newblock Reinforcement learning with augmented data, 2020.

\bibitem[{Lee} et~al.(2019){Lee}, {Nagabandi}, {Abbeel}, and
  {Levine}]{lee2019slac}
A.~X. {Lee}, A.~{Nagabandi}, P.~{Abbeel}, and S.~{Levine}.
\newblock Stochastic latent actor-critic: Deep reinforcement learning with a
  latent variable model.
\newblock \emph{arXiv e-prints}, 2019.

\bibitem[Lee et~al.(2019)Lee, Nagabandi, Abbeel, and Levine]{lee2019stochastic}
Alex~X Lee, Anusha Nagabandi, Pieter Abbeel, and Sergey Levine.
\newblock Stochastic latent actor-critic: Deep reinforcement learning with a
  latent variable model.
\newblock \emph{arXiv preprint arXiv:1907.00953}, 2019.

\bibitem[Lillicrap et~al.(2015{\natexlab{a}})Lillicrap, Hunt, Pritzel, Heess,
  Erez, Tassa, Silver, and Wierstra]{lillicrap2015ddpg}
Timothy~P. Lillicrap, Jonathan~J. Hunt, Alexander Pritzel, Nicolas Heess, Tom
  Erez, Yuval Tassa, David Silver, and Daan Wierstra.
\newblock Continuous control with deep reinforcement learning.
\newblock \emph{CoRR}, 2015{\natexlab{a}}.

\bibitem[Lillicrap et~al.(2015{\natexlab{b}})Lillicrap, Hunt, Pritzel, Heess,
  Erez, Tassa, Silver, and Wierstra]{lillicrap2015continuous}
Timothy~P. Lillicrap, Jonathan~J. Hunt, Alexander Pritzel, Nicolas Manfred~Otto
  Heess, Tom Erez, Yuval Tassa, David Silver, and Daan Wierstra.
\newblock Continuous control with deep reinforcement learning.
\newblock \emph{CoRR}, abs/1509.02971, 2015{\natexlab{b}}.

\bibitem[Mnih et~al.(2013)Mnih, Kavukcuoglu, Silver, Graves, Antonoglou,
  Wierstra, and Riedmiller]{mnih2013dqn}
Volodymyr Mnih, Koray Kavukcuoglu, David Silver, Alex Graves, Ioannis
  Antonoglou, Daan Wierstra, and Martin Riedmiller.
\newblock Playing atari with deep reinforcement learning.
\newblock \emph{arXiv e-prints}, 2013.

\bibitem[Mnih et~al.(2016{\natexlab{a}})Mnih, Badia, Mirza, Graves, Lillicrap,
  Harley, Silver, and Kavukcuoglu]{mnih2016a3c}
Volodymyr Mnih, Adria~Puigdomenech Badia, Mehdi Mirza, Alex Graves, Timothy~P.
  Lillicrap, Tim Harley, David Silver, and Koray Kavukcuoglu.
\newblock Asynchronous methods for deep reinforcement learning.
\newblock \emph{CoRR}, 2016{\natexlab{a}}.

\bibitem[Mnih et~al.(2016{\natexlab{b}})Mnih, Badia, Mirza, Graves, Lillicrap,
  Harley, Silver, and Kavukcuoglu]{mnih2016asynchronous}
Volodymyr Mnih, Adrià~Puigdomènech Badia, Mehdi Mirza, Alex Graves,
  Timothy~P. Lillicrap, Tim Harley, David Silver, and Koray Kavukcuoglu.
\newblock Asynchronous methods for deep reinforcement learning,
  2016{\natexlab{b}}.

\bibitem[Munos et~al.(2016)Munos, Stepleton, Harutyunyan, and
  Bellemare]{munos2016retrace}
R{\'{e}}mi Munos, Tom Stepleton, Anna Harutyunyan, and Marc~G. Bellemare.
\newblock Safe and efficient off-policy reinforcement learning.
\newblock \emph{CoRR}, 2016.

\bibitem[Noroozi and Favaro(2016)]{noroozi2016unsupervised}
Mehdi Noroozi and Paolo Favaro.
\newblock Unsupervised learning of visual representations by solving jigsaw
  puzzles.
\newblock In \emph{European Conference on Computer Vision}, pages 69--84.
  Springer, 2016.

\bibitem[Pinto et~al.(2016)Pinto, Gandhi, Han, Park, and
  Gupta]{pinto2016curious}
Lerrel Pinto, Dhiraj Gandhi, Yuanfeng Han, Yong{-}Lae Park, and Abhinav Gupta.
\newblock The curious robot: Learning visual representations via physical
  interactions.
\newblock \emph{CoRR}, 2016.

\bibitem[Raileanu et~al.(2020)Raileanu, Goldstein, Yarats, Kostrikov, and
  Fergus]{raileanu2020automatic}
Roberta Raileanu, Max Goldstein, Denis Yarats, Ilya Kostrikov, and Rob Fergus.
\newblock Automatic data augmentation for generalization in deep reinforcement
  learning.
\newblock \emph{CoRR}, 2020.

\bibitem[Schwarzer et~al.(2020{\natexlab{a}})Schwarzer, Anand, Goel, Hjelm,
  Courville, and Bachman]{schwarzer2020data}
Max Schwarzer, Ankesh Anand, Rishab Goel, R~Devon Hjelm, Aaron Courville, and
  Philip Bachman.
\newblock Data-efficient reinforcement learning with momentum predictive
  representations.
\newblock \emph{arXiv preprint arXiv:2007.05929}, 2020{\natexlab{a}}.

\bibitem[Schwarzer et~al.(2020{\natexlab{b}})Schwarzer, Anand, Goel, Hjelm,
  Courville, and Bachman]{schwarzer2020spr}
Max Schwarzer, Ankesh Anand, Rishab Goel, R.~Devon Hjelm, Aaron~C. Courville,
  and Philip Bachman.
\newblock Data-efficient reinforcement learning with momentum predictive
  representations.
\newblock \emph{CoRR}, 2020{\natexlab{b}}.

\bibitem[Silver et~al.(2014)Silver, Lever, Heess, Degris, Wierstra, and
  Riedmiller]{silver14dpg}
David Silver, Guy Lever, Nicolas Heess, Thomas Degris, Daan Wierstra, and
  Martin Riedmiller.
\newblock Deterministic policy gradient algorithms.
\newblock In \emph{Proceedings of the 31st International Conference on Machine
  Learning}, 2014.

\bibitem[Srinivas et~al.(2020)Srinivas, Laskin, and Abbeel]{srinivas2020curl}
Aravind Srinivas, Michael Laskin, and Pieter Abbeel.
\newblock Curl: Contrastive unsupervised representations for reinforcement
  learning.
\newblock \emph{arXiv preprint arXiv:2004.04136}, 2020.

\bibitem[Stooke et~al.(2020)Stooke, Lee, Abbeel, and
  Laskin]{stooke2020decoupling}
Adam Stooke, Kimin Lee, Pieter Abbeel, and Michael Laskin.
\newblock Decoupling representation learning from reinforcement learning.
\newblock \emph{arXiv preprint arXiv}, 2020.

\bibitem[Tassa et~al.(2012)Tassa, Erez, and Todorov]{tassa2012trajopt}
Yuval Tassa, Tom Erez, and Emanuel Todorov.
\newblock Synthesis and stabilization of complex behaviors through online
  trajectory optimization.
\newblock In \emph{2012 IEEE/RSJ International Conference on Intelligent Robots
  and Systems}, 2012.

\bibitem[Tassa et~al.(2018)Tassa, Doron, Muldal, Erez, Li, Casas, Budden,
  Abdolmaleki, Merel, Lefrancq, et~al.]{tassa2018dmcontrol}
Yuval Tassa, Yotam Doron, Alistair Muldal, Tom Erez, Yazhe Li, Diego de~Las
  Casas, David Budden, Abbas Abdolmaleki, Josh Merel, Andrew Lefrancq, et~al.
\newblock Deepmind control suite.
\newblock \emph{arXiv preprint arXiv:1801.00690}, 2018.

\bibitem[Todorov et~al.(2012)Todorov, Erez, and Tassa]{todorov2012mujoco}
Emanuel Todorov, Tom Erez, and Yuval Tassa.
\newblock Mujoco: A physics engine for model-based control.
\newblock In \emph{2012 IEEE/RSJ International Conference on Intelligent Robots
  and Systems}, 2012.

\bibitem[Vincent et~al.(2008)Vincent, Larochelle, Bengio, and
  Manzagol]{vincent2008extracting}
Pascal Vincent, Hugo Larochelle, Yoshua Bengio, and Pierre-Antoine Manzagol.
\newblock Extracting and composing robust features with denoising autoencoders.
\newblock In \emph{Proceedings of the 25th international conference on Machine
  learning}, pages 1096--1103. ACM, 2008.

\bibitem[Wang and Gupta(2015)]{Wang_UnsupICCV2015}
Xiaolong Wang and Abhinav Gupta.
\newblock Unsupervised learning of visual representations using videos.
\newblock In \emph{ICCV}, 2015.

\bibitem[Watkins and Dayan(1992)]{watkins1992qlearning}
Christopher J. C.~H. Watkins and Peter Dayan.
\newblock Q-learning.
\newblock \emph{Machine Learning}, 1992.

\bibitem[Watkins(1989)]{watkins1989phd}
Christopher John Cornish~Hellaby Watkins.
\newblock \emph{Learning from Delayed Rewards}.
\newblock PhD thesis, King's College, 1989.

\bibitem[Yan et~al.(2020)Yan, Vangipuram, Abbeel, and Pinto]{yan2020learning}
Wilson Yan, Ashwin Vangipuram, Pieter Abbeel, and Lerrel Pinto.
\newblock Learning predictive representations for deformable objects using
  contrastive estimation.
\newblock \emph{arXiv preprint arXiv:2003.05436}, 2020.

\bibitem[Yarats et~al.(2019)Yarats, Zhang, Kostrikov, Amos, Pineau, and
  Fergus]{yarats2019improving}
Denis Yarats, Amy Zhang, Ilya Kostrikov, Brandon Amos, Joelle Pineau, and Rob
  Fergus.
\newblock Improving sample efficiency in model-free reinforcement learning from
  images.
\newblock \emph{arXiv preprint arXiv:1910.01741}, 2019.

\bibitem[Yarats et~al.(2021{\natexlab{a}})Yarats, Fergus, Lazaric, and
  Pinto]{yarats2021proto}
Denis Yarats, Rob Fergus, Alessandro Lazaric, and Lerrel Pinto.
\newblock Reinforcement learning with prototypical representations.
\newblock \emph{CoRR}, 2021{\natexlab{a}}.

\bibitem[Yarats et~al.(2021{\natexlab{b}})Yarats, Kostrikov, and
  Fergus]{yarats2021image}
Denis Yarats, Ilya Kostrikov, and Rob Fergus.
\newblock Image augmentation is all you need: Regularizing deep reinforcement
  learning from pixels.
\newblock In \emph{9th International Conference on Learning Representations,
  ICLR 2021}, 2021{\natexlab{b}}.

\bibitem[Young et~al.(2020)Young, Gandhi, Tulsiani, Gupta, Abbeel, and
  Pinto]{young2020visual}
Sarah Young, Dhiraj Gandhi, Shubham Tulsiani, Abhinav Gupta, Pieter Abbeel, and
  Lerrel Pinto.
\newblock Visual imitation made easy.
\newblock \emph{CoRR}, 2020.

\bibitem[Zhan et~al.(2020)Zhan, Zhao, Pinto, Abbeel, and
  Laskin]{zhan2020framework}
Albert Zhan, Philip Zhao, Lerrel Pinto, Pieter Abbeel, and Michael Laskin.
\newblock A framework for efficient robotic manipulation.
\newblock \emph{CoRR}, 2020.

\bibitem[Zhang et~al.(2017)Zhang, Isola, and Efros]{zhang2017split}
Richard Zhang, Phillip Isola, and Alexei~A Efros.
\newblock Split-brain autoencoders: Unsupervised learning by cross-channel
  prediction.
\newblock In \emph{Proceedings of the IEEE Conference on Computer Vision and
  Pattern Recognition}, pages 1058--1067, 2017.

\end{thebibliography}
\bibliographystyle{plainnat}

\includeappendixtrue % comment it out to hide appendix

\ifincludeappendix

\appendix
\section*{Appendix}

\section{Benchmarks}
\label{section:benchmarks}
We classify a set of $24$ continuous control tasks from DMC~\citep{tassa2018dmcontrol} into \textit{easy}, \textit{medium}, and \textit{hard} benchmarks and provide a summary for each task in~\cref{table:benchamrks}.

\begin{table}[!h]
\centering
\begin{tabular}{lccccc}
\hline
Task & Traits &  Difficulty & Allowed Steps &$\mathrm{dim}(\gS)$ & $\mathrm{dim}(\gA)$   \\

\hline
Cartpole Balance & balance, dense & easy & $1\times 10^6$  & $4$ & $1$  \\
Cartpole Balance Sparse & balance, sparse & easy & $1\times 10^6$  & $4$ & $1$  \\
Cartpole Swingup & swing & dense & $1\times 10^6$  & $4$ & $1$ \\
Cup Catch & swing, catch, sparse & easy & $1\times 10^6$  & $8$ & $2$ \\
Finger Spin & rotate, dense & easy & $1\times 10^6$  & $6$ & $2$ \\
Hopper Stand & stand, dense & easy & $1\times 10^6$  & $14$ & $4$ \\
Pendulum Swingup & swing, sparse & easy & $1\times 10^6$  & $2$ & $1$ \\
Walker Stand  & stand, dense & easy & $1\times 10^6$  & $18$ & $6$\\
Walker Walk  & walk, dense & easy & $1\times 10^6$  & $18$ & $6$ \\
\hline 
Acrobot Swingup & diff. balance, dense & medium & $3\times 10^6$  & $4$ & $1$  \\
Cartpole Swingup Sparse & swing, sparse & medium & $3\times 10^6$  & $4$ & $1$  \\
Cheetah Run & run, dense & medium & $3\times 10^6$  & $18$ & $6$  \\
Finger Turn Easy & turn, sparse & medium & $3\times 10^6$  & $6$ & $2$  \\
Finger Turn Hard & turn, sparse & medium & $3\times 10^6$  & $6$ & $2$  \\
Hopper Hop & move, dense & medium & $3\times 10^6$  & $14$ & $4$  \\
Quadruped Run & run, dense & medium & $3\times 10^6$  & $56$ & $12$  \\
Quadruped Walk & walk, dense & medium & $3\times 10^6$  & $56$ & $12$  \\
Reach Duplo & manipulation, sparse & medium & $3\times 10^6$  & $55$ & $9$  \\
Reacher Easy & reach, dense & medium & $3\times 10^6$  & $4$ & $2$  \\
Reacher Hard & reach, dense & medium & $3\times 10^6$  & $4$ & $2$  \\
Walker Run & run, dense & medium & $3\times 10^6$  & $18$ & $6$  \\
\hline
Humanoid Stand & stand, dense & hard & $30\times 10^6$  & $54$ & $21$  \\
Humanoid Walk & walk, dense & hard & $30\times 10^6$  & $54$ & $21$  \\
Humanoid Run & run, dense & hard & $30\times 10^6$  & $54$ & $21$  \\

\hline
\end{tabular}
\caption{\label{table:benchamrks} A detailed description of each tasks in our \textit{easy}, \textit{medium}, and \textit{hard} benchmarks.}
\end{table}

\newpage

\section{Hyper-parameters}
\label{section:hyperparams}

The full list of hyper-parameters is presented in~\cref{table:hyper_params}. While we tried to keep the settings identical for each of the task, there are a few specific deviations for some tasks.\\
\textbf{Walker Stand/Walk/Run} For all three  tasks we use mini-batch size of $512$ and $n$-step return of $1$.\\
\textbf{Quadruper Run} We set the replay buffer size to $10^5$.\\
\textbf{Humanoid Stand/Walk} We set learning rate to $8\times 10^{-5}$ and increase features dim. to $100$.

\begin{table}[hb!]
\caption{\label{table:hyper_params} A default set of hyper-parameters used in our experiments.}
\centering
\begin{tabular}{lc}
\hline
Parameter        & Setting \\
\hline
Replay buffer capacity & $10^6$ \\
Action repeat & $2$ \\
Seed frames & $4000$ \\
Exploration steps & $2000$ \\
$n$-step returns & $3$ \\
Mini-batch size & $256$ \\
Discount $\gamma$ & $0.99$ \\
Optimizer & Adam \\
Learning rate & $10^{-4}$ \\
Agent update frequency & $2$ \\
Critic Q-function soft-update rate $\tau$ & $0.01$ \\
Features dim. & $50$ \\
Hidden dim. & $1024$ \\
Exploration stddev. clip & $0.3$ \\
\multirow{3}{*}{Exploration stddev. schedule}& easy: $\mathrm{linear}(1.0, 0.1, 100000)$\\ & medium: $\mathrm{linear}(1.0, 0.1, 500000)$\\& hard: $\mathrm{linear}(1.0, 0.1, 2000000)$\\

\hline
\end{tabular}

\end{table}

\fi

\end{document}